\documentclass[
]{ceurart}

\usepackage{listings}
\begin{document}

\copyrightyear{2022}
\copyrightclause{Copyright for this paper by its authors.
  Use permitted under Creative Commons License Attribution 4.0
  International (CC BY 4.0).}

\conference{Woodstock'22: Symposium on the irreproducible science,
  June 07--11, 2022, Woodstock, NY}


\title{Enhancing Programming eTextbooks with ChatGPT Generated Counterfactual-Thinking-Inspired Questions}

\tnotemark[1]
\tnotetext[1]{You can use this document as the template for preparing your
  publication. We recommend using the latest version of the ceurart style.}


\author[1]{Arun Balajiee Lekshmi-Narayanan}[%
orcid=0000-0002-7735-5008,
email=arl122@pitt.edu,
url=https://a2un.github.io,
]
\cormark[1]
\fnmark[1]

\author[1]{Rully Agus Hendrawan}[%
orcid=0000-0001-8541-0305,
email=ruhendrawan@pitt.edu,
url=https://ruhendrawan.com/,
]
\cormark[1]
\fnmark[1]




\author[2]{Venktesh V}[%
email=venktesh.v@iiitd.ac.in
]

\address[1]{University of Pittsburgh, Pittsburgh, PA, USA}
\address[2]{Indraprastha Institute of Information Technology, Delhi,India}

\cortext[1]{Corresponding author.}
\fntext[1]{These authors contributed equally.}

\begin{abstract}
Digital textbooks have become an integral part of everyday learning tasks. In this work, we consider the use of digital textbooks for programming classes. Generally, students struggle with utilizing textbooks on programming to the maximum, with a possible reason being that the example programs provided as illustration of concepts in these textbooks don't offer sufficient interactivity for students, and thereby not sufficiently motivating to explore or understand these programming examples better. In our work, we explore the idea of enhancing the navigability of intelligent textbooks with the use of ``counterfactual'' questions, to make students think critically about these programs and enhance possible program comprehension. Inspired from previous works on nudging students on counter factual thinking, we present the possibility to enhance digital textbooks with questions generated using GPT--3.5.
\end{abstract}


\maketitle

\section{Introduction \& Related Work}
Interactive textbooks have been explored extensively among the computer science education community, for many years~\cite{rossling2006merging}, with the best examples of these being ELM-ART~\cite{weber2001elmart} for LISP programming; OpenDSA~\cite{Fouh2014} an online reading platform for topics in Data Structures and Algorithms;  Runestone~\cite{Ericson2020} a platform to host interactive programming practice material, among other, suggesting that some of the most common eTextbooks for computer science education are on programming. In this work, we consider the enhancement of digital textbooks with the use of newer tools such as Large Language Models, as explored in the recent efforts on the use of Large Language Models to generate questions for intelligent textbooks~\cite{openStaxQG}.

From prior work, we have looked at cases of support for program comprehension using question generation. Specifically, these approaches are directed at some of the following ideas, namely, 

\begin{enumerate}
  \item \textbf{Critical Thinking and Perspective Taking}: Promote a growth mindset \cite{manniam2018virtual} and foster robust critical thinking that nudges students towards a higher understanding. This might help students take a step back \cite{10.1145/3586030}, and think slower and more deeply, encouraging perspective-taking.

  \item \textbf{Syntax Analysis}: Encourage students to think based on specific errors and functional keywords within programs \cite{santos2022jask} about the syntax of the programs, deepening their understanding of syntactic elements.

  \item \textbf{Goal-Oriented Analysis}: Focus on the steps and possibilities when tracing a program \cite{russell2022automated, stankov2023smart}, which is essentially about understanding the objective or the goal of the code and how it is achieved. Their work encourages students to think about what changes in the program would mean for its final output or goal.

  \item \textbf{Problem-Solution Mapping}: Generate questions based on misconceptions \cite{henley2021inquisitive}. The approach of asking why or why not a certain line in a program works encourages students to think about the problem that the specific line of code is solving. This helps them reverse engineer the code, i.e., map the solution (code) to the problem.

  \item \textbf{Intrinsic Program Analysis}: Encourage students to deeply analyze the program by generating questions that are intrinsically tied to the program \cite{tamang2022automatic}, while stimulating students to think about the program and its inherent aspects \cite{lehtinen2023automated}.
    
    \item \textbf{Large Language Model Prompt-based QG}: While the focus of the question generation using LLM for intelligent digital textbooks in prior work~\cite{openStaxQG} is similar to the current work, our work attempts to address the notion of prompting the LLM to an extent where the questions generated could be hierarchical, also specifically, counterfactual questions that allow room for alternative and creative mental constructs that support program comprehension.

\end{enumerate}

In most cases, the question generation is not the key idea to enhance program comprehension. And in the cases where the program comprehension is supported by question generation ~\cite{tamang2022automatic,lehtinen2023automated}, the approaches have specific types of questions that they generate using fixed templates for the generation tasks using ``in-filling'' approach.

Additionally, Lehtinen and colleagues~\cite{lehtinen2023automated} construct the question types from prior framework, but these ideas do not necessarily consider an hierarchical approach towards a step-wise program comprehension. If we refer to the past work on program comprehension~\cite{Schulte2008BlockMA}, we can proposition that questions that invoke students to think critically can also be hierarchical. Some other ideas that support this are the ICAP Framework~\cite{Chi2014TheIF}







Hence, our contributions in this work are three-fold:

\begin{enumerate}
    \item \textit{Systematically explore the use of prompt mechanism in the latest freely available ChatGPT (GPT--3.5) system to generate questions for a JAVA Program}
    \item \textit{Understand the relations between the ChatGPT (GPT--3.5) suggested categorization of these questions and those annotated by humans. Such an understanding could promote transparency in the process of LLM response to prompts.}
    \item \textit{Present a dataset of questions that are at varying levels in the hierarchy used to describe the program comprehension task. These questions could act as an enhancement to an existing intelligent, digital textbook on programming}
\end{enumerate}

\section{Counterfactual-Thinking Inspired Questions Generation}

Building on existing coding challenges, we generate a set of questions inspired by previous works related to counterfactual thinking. These questions are then manually categorized into distinct themes, enabling a deeper understanding of prevalent trends and insights into the application of counterfactual thinking in problem-solving and creative tasks within programming.

\subsection{Code Challenges}

Practice exercises help novice programmers learn how to code. The process of learning to code involves understanding various programming concepts and practicing them until they become second nature. This is where practice exercises play a crucial role. These exercises provide an opportunity for novice programmers to apply the theory they have learned, strengthen their problem-solving skills, and learn from trial and error. Sample code challenges are categorized according to their functionality as follows:
\begin{enumerate}
  \item Object and Arithmetic: Student Profile, Circle Area Calculator, Coordinate Shift, Shape Measurements
  \item Repeated Calculation: Average Calculator, Bingo Board, Grade Calculator, Multiplication Table, Prime Checker
  \item Comparisons and Rules: Place Name Comparator, Age Comparison, Phone Buyer, Bank Account
\end{enumerate}

We generated these programs using ChatGPT with the goal descriptions provided in DeepCodeDataset~\cite{Rus2022DeepCodeAA}. The prompts to ChatGPT were so it generates the program as a novice programmer would. 

\subsection{Question Generation with LLM}

The efficacy of Language Learning Models (LLMs) is noteworthy \cite{webson2023language}. If prompted correctly, an LLM can assist teachers in generating programming questions. We develop 5 prompts to generate questions for each code challenge. Every Prompt starts with the phrase \textit{``In tabular file format with the following columns (LineNumber, LineCode, Question) generate counterfactual questions to make students critically think about the program from .. ''}, except for a change with a few keywords as specified in Table~\ref{tab:prompts}.

\begin{table}[t!]
    \centering
    \begin{tabular}{|c|c|} \hline
       \textbf{Category}  & \textbf{Prompt Keywords} \\ \hline
         & \\ 
         Critical Thinking and Perspective Taking & \textit{different perspectives or aspects} \\
         & \\ 
         Syntax Analysis & \textit{syntactic perspectives or aspects} \\
         & \\ 
         Goal-Oriented Analysis & \textit{such that final goal of} \\
         & \textit{the program is changed} \\
         & \\ 
         Problem-Solution Mapping & \textit{could ask that have} \\
         & \textit{the same program as the solution}\\
         & \\ 
         Intrinsic Program Analysis &  \textit{based on the program and the answer to }\\
         & \textit{those questions lie within the program}\\
         \bottomrule
    \end{tabular}
    \caption{List of various Prompts  Keywords used to Generate questions of a certain category}
    \label{tab:prompts}
\end{table}

\subsection{Labeling Generated Questions}

The classes used to label questions asked about a piece of code or a programming concept are:

\begin{enumerate}
    \item \textbf{Syntax (S)}: The ability of code to compile successfully depends on whether it adheres to the syntax rules of the programming language. The concept of syntax in a programming language includes the proper use of brackets, punctuation like semicolons or colons, variable declaration, and so on. E.g., \textit{"What would happen if we forgot to include the semicolon at the end?"}

    \item \textbf{Programming Logic (PL)}: Logical comprehension and overall understanding of a piece of code or a program. The emphasis is not on specific syntax or language constructs, but rather on understanding how the pieces of the code fit together to create a functioning program. This could involve understanding the purpose of specific variables or functions, how control flow structures like loops or conditionals are used, or the logic behind a specific algorithm or data structure used in the code. In some cases, these questions could involve modifications or adaptations of existing code to meet new requirements or goals. E.g. \textit{"What if we needed to read input from a file instead of from the user? How could we modify the code to achieve this?"}

    \item \textbf{Goal-oriented (G)}: Achieving a desired result regardless of the specifics of the code. This category is more focused on the problem-solving aspect of programming, where the specific language or implementation details are secondary to the overall objective. E.g. \textit{"What if we wanted to read the radius value from a file instead of user input?"}

    \item \textbf{Miscellaneous (M)}: A catch-all category for questions that don't neatly fit into the other categories. This could involve questions about programming best practices, questions about specific development tools or environments, version control, debugging strategies, or questions about the broader principles and philosophies of software development. E.g., \textit{"Why is the main method necessary in a Java program?"}
\end{enumerate}

\subsection{Thematic Categorization of Generated Questions}

Once the set of questions is generated, the next step is to categorize them. The categorization phase is an iterative process that seeks to classify the generated questions based on distinct themes. This grouping process is designed to be flexible and iterative, allowing for the continuous refinement of categories as manifested themes or nuances are emerging during the analysis process. We focus on themes that are specific to programming tasks rather than broader aspects of problem-solving or creativity.

Each question is analyzed focused on its intent, the concepts it explores, and the specific aspects of programming it addresses. The emergent themes arise organically from the content of the questions, which can cover a diverse range of topics. The end result of this categorization phase is a collection of thematic categories, as reflected in the questions it comprises. Each theme represents a unique challenge that students are facing in learning to program with the application of counterfactual thinking.


\section{Discussions}

\subsection{Themes}

\begin{figure}[h!]
    \centering
    \includegraphics[width=\textwidth]{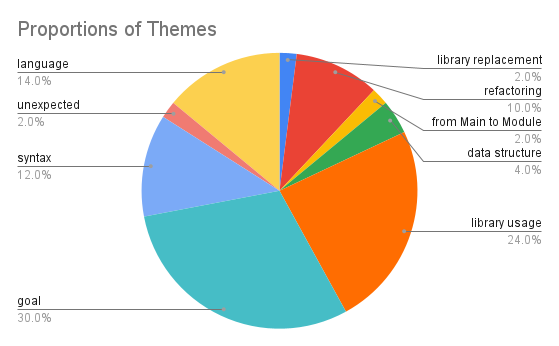}
    \caption{With thematic analysis provides a perspective on the question types}
    \label{fig:themes-prop}
\end{figure}

The themes that we found on the generated questions are:

\begin{enumerate}
    \item \textbf{Language Understanding - Syntax}: Pertain to the syntax and the grammatical rules of the programming language. E.g., \textit{"Why is there a semicolon at the end of this line in a Java program?"}.

    \item \textbf{Language Understanding - Semantic}: Relate to the meaning or purpose of specific programming language keywords or constructs, as well as the principles behind how they operate.  E.g., \textit{"What does the 'final' keyword do in Java?"}.

    \item \textbf{Language Understanding - Other}: Grasping the core concepts, conventions, and idiosyncrasies of a specific programming language. This theme could involve a broad range of language features. E.g., \textit{"Why is the main method necessary in a Java program?"}

    \item \textbf{Library/Function Understanding}: Understanding how a specific library, module, or function works. This could be a standard library in a language or a third-party library. E.g., \textit{"What does the 'Scanner' class do?"}.

    \item \textbf{External Behaviour}: How a program interacts with things outside of itself (like input/output operations, network requests, etc.) or how it manifests its results. This could involve its inputs or outputs, how it handles data from a user or a file, or the visible effects of its operation (like changes to the UI or the production of log messages). E.g., \textit{"What will this program print?"} or \textit{"What happens when I click this button?"}.

    \item \textbf{Refactoring, Internal Behaviour}: How the internal logic of the code could be changed without affecting its external behavior. This could be modified, optimized, or restructured without changing what it does. They can also involve understanding the internal workings of a piece of code, such as the control flow or the interactions between different components of a program. E.g., \textit{"What would happen if we replaced this loop with a map function?"}.

\end{enumerate}

We have summarized our proportions in Figure~\ref{fig:themes-prop}, suggests a richer thematic analysis provides us with sufficient information. This could indicate that the number of codes developed by us may be incomplete, and could be improved.

\section{Conclusion \& Future Work}

While there are several directions to consider with our current work, one potential direction that we are likely going to continue explore is with their use as aids in generating smart content~\cite{smart-content}.

\subsection{Fine-Grained Navigation Support in Intelligent Textbooks}


Intelligent textbooks are transforming education \cite{intelligenttextbooks}, offering dynamic and interactive learning experiences. Navigation within these textbooks is a crucial aspect that can enhance learning efficiency and engagement. This study demonstrates how the thematic classification of questions, inspired by counterfactual thinking, and generated by pre--trained Large Language Models (LLMs), can enrich navigational support in intelligent textbooks.

Counterfactual thinking, a mental simulation of alternative realities, often prompted by "what if" scenarios. By combining this psychological construct with the capabilities of pre-trained LLMs, we propose an approach to generate coding-related questions, potentially improving programming proficiency. In our work, even if we do not provide sufficient evidence to the efficacy of the use of LLMs for this task, it indicates small successes in the use of LLMs to generate questions suitable to promote and support program comprehension.

Question themes represent a critical area in computer science education, enhancing learners' problem-solving abilities, coding efficiency, and overall understanding of programming languages. These thematic classifications facilitate learners in personalizing their learning journey and allow for more focused and efficient navigation through the material in intelligent textbooks.





\begin{acknowledgments}
Many thanks to Khushboo Thaker and Arezou Farzaneh for their inputs with this work without which this work would not have been complete with.
\end{acknowledgments}

\bibliography{main}

\appendix

\section{Online Resources}

All the code and resources are avaiable via \href{https://github.com/PAWSLabUniversityOfPittsburgh/ChatGPTNovice/}{Private Github Repository}

\end{document}